\documentclass{article}

\usepackage[numbers]{natbib}

\usepackage[final]{neurips_2022}

\usepackage[utf8]{inputenc} % allow utf-8 input
\usepackage[T1]{fontenc}    % use 8-bit T1 fonts
\usepackage{hyperref}       % hyperlinks
\usepackage{url}            % simple URL typesetting
\usepackage{booktabs}       % professional-quality tables
\usepackage{amsfonts}       % blackboard math symbols
\usepackage{nicefrac}       % compact symbols for 1/2, etc.
\usepackage{microtype}      % microtypography
\usepackage{graphicx}
\usepackage{subcaption}
\usepackage{multirow}
\usepackage[dvipsnames]{xcolor}
\usepackage{amsmath}
\usepackage{changepage}
\usepackage[ruled,vlined]{algorithm2e}
\usepackage[toc,page,header]{appendix}
\usepackage{minitoc}
\usepackage{enumitem}
\setlist{nolistsep}
\DeclareMathOperator*{\argmin}{arg\,min}
\DeclareMathOperator*{\argmax}{arg\,max}

\newcommand{\C}{\mathcal{C}}
\newcommand{\beq}{\begin{equation}}
\newcommand{\eeq}{\end{equation}}
\newcommand{\beqnn}{\begin{equation*}}
\newcommand{\eeqnn}{\end{equation*}}

\newcommand{\MLP}{\textrm{MLP}}
\newcommand{\MHA}{\textrm{MHA}}
\newcommand{\bx}{\boldsymbol{x}}
\newcommand{\bh}{\boldsymbol{h}}
\newcommand{\bbias}{\boldsymbol{b}}
\newcommand{\Nl}{\mathcal{N}_\ell}
\newcommand{\N}{\mathcal{N}}

\title{Neural Topological Ordering for Computation Graphs}

\author{%
Mukul Gagrani\thanks{Equal contribution} \\
Qualcomm AI Research\thanks{Qualcomm AI Research is an initiative of Qualcomm Technologies, Inc.}\\
\texttt{mgagrani@qti.qualcomm.com} \\
\And
Corrado Rainone$^*$ \\
Qualcomm AI Research\\
\texttt{crainone@qti.qualcomm.com} \\
\And
Yang Yang\thanks{Work completed during employment at Qualcomm Technologies, Inc.} \\
Google LLC\\
\And
Harris Teague \\
Qualcomm AI Research\\
\And
Wonseok Jeon \\
Qualcomm AI Research\\
\And
Herke Van Hoof \\
University of Amsterdam, 
Netherlands\\
\And
Weiliang Will Zeng \\
Qualcomm AI Research\\
\And
Piero Zappi\\
Qualcomm AI Research\\
\AND
Christopher Lott \\
Qualcomm AI Research\\
\And
Roberto Bondesan \\
Qualcomm AI Research\\
}

\begin{document}

\maketitle

\begin{abstract}
%   The abstract paragraph should be indented \nicefrac{1}{2}~inch (3~picas) on
%   both the left- and right-hand margins. Use 10~point type, with a vertical
%   spacing (leading) of 11~points.  The word \textbf{Abstract} must be centered,
%   bold, and in point size 12. Two line spaces precede the abstract. The abstract
%   must be limited to one paragraph.
    Recent works on machine learning for combinatorial optimization have shown that learning based approaches can outperform heuristic methods in terms of speed and performance. 
    In this paper, we consider the problem of finding an optimal topological order on a directed acyclic graph with focus on the memory minimization problem which arises in compilers. We propose an end-to-end machine learning based approach for topological ordering using an encoder-decoder framework. Our encoder is a novel attention based graph neural network architecture called \emph{Topoformer} which uses different topological transforms of a DAG for message passing. The node embeddings produced by the encoder are converted into node priorities which are used by the decoder to generate a probability distribution over topological orders. We train our model on a dataset of synthetically generated graphs called layered graphs. We show that our model outperforms, or is on-par, with several topological ordering baselines while being significantly faster on synthetic graphs with up to 2k nodes. We also train and test our model on a set of real-world computation graphs, showing performance improvements.

\end{abstract}

\section{Introduction \label{sec:intro}}

% \mukul{comment}
% \corrado{comment}
% \yang{comment}
% \wonseok{comment}
% \harris{comment}
% \roberto{comment}
% \will{comment}
% \piero{comment}
% \herke{comment}

% \newcommand{\mukul}[1]{{\color{red}MG: #1}}
% \newcommand{\corrado}[1]{{\color{BlueGreen}CR: #1}}
% \newcommand{\yang}[1]{{\color{Salmon}YY: #1}}
% \newcommand{\wonseok}[1]{{\color{cyan}WJ: #1}}
% \newcommand{\harris}[1]{{\color{lime}HT: #1}}
% \newcommand{\roberto}[1]{{\color{orange}RB: #1}}
% \newcommand{\will}[1]{{\color{pink}WZ: #1}}
% \newcommand{\piero}[1]{{\color{lightgray}PZ: #1}}
% \newcommand{\herke}[1]{{\color{olive}HH: #1}}

Many problems in computer science amount to finding the best sequence of objects consistent with some precedence constraints.
An intuitive example comes from routing problems, where we would like to find the shortest route between cities but we have requirements (i.e. for example to pick up and subsequently deliver a package) on the order in which the cities should be visited~\cite{vrp_book}. Another case is found in compiler pipelines, wherein the "cities" become operations to be executed and the constraints come from the data dependencies between these operations, such as when the result of an operation is an operand in a subsequent one. In this case, the metric to be optimized can be the run time of the compiled program, or the memory required to execute the program \cite{ahn2020ordering}.
Common across this class of problems is their formulation in term of finding the optimal topological order of the Directed Acyclic Graph (DAG) that encodes the precedence constraints, which induces a Combinatorial Optimization~\cite{co_book} (CO) problem which is in general computationally hard \cite{sharp_p_complete}.

Already from the two examples above, one can immediately grasp the relevance of such problems for industrial Operations Research, which has prompted various actors to invest in the development of efficient CO solvers; these solvers usually encapsulate heuristic methods whose design typically requires extensive use of domain-specific and problem-specific knowledge, across decades of development. In recent years, considerable interest has emerged in the possibility of replacing such handcrafted heuristics with ones learned by deep neural nets~\cite{bengio2021machine} (machine learning for combinatorial optimization, MLCO). As a matter of fact, both of our two examples of DAG-based CO problems have indirectly been object of study in the Machine Learning literature. References~\cite{kool2018TSP,xin2021neurolkh,joshi2021learning,correia2022neural} take into consideration Routing Problems, especially the Traveling Salesperson Problem (TSP) which, on account of its richness, complexity and long history of mathematical study~\cite{tsp_book}, has attained the status of a standard benchmark for MLCO~\cite{joshi2021learning}. Conversely, less attention has been devoted to operations sequencing likely due to the proprietary and sensitive nature of compiler workflows, which hampers the definition of public benchmarks. References~\cite{regalpaper, gopaper} both consider the task of optimizing the run time of a neural network's forward pass by optimizing the ordering and device assignment of its required operations. However, in this last case the sequencing stage is only one part of a larger compiler pipeline, and as a result of this both the performance metrics and the datasets employed cannot be made available for reproduction by third parties. This makes it both hard to assess the results therein, and to draw general conclusions and guidelines for the advancement of MLCO, which still suffers from a lack of commonly accepted and standard datasets and benchmarks.

In this work, we address the problem of finding optimal topological orders in a DAG using deep learning, focusing on the compiler task of optimizing the peak local memory usage during execution. We make the following contributions:
%. We focus on a compiler setting wherein the metric to optimize is the Peak Memory Usage during execution, and do so while minding the issue of reproducibility: our performance metric is transparently defined, the baselines with which we compare are classic and widely available algorithms for graph sequencing, and we introduce an algorithm to generate a synthetic dataset of computation graphs for the training, validation, and testing of generic ML architectures for sequencing; we report compelling results, showing that our Neural method is able to match, on a large and reproducible set of graphs, the performance of classic baselines that require as much as 1000x more run time.

\begin{itemize}[leftmargin=4.2mm]
    \item We present a neural framework to optimize sequences on directed acyclic graphs. Mindful of the need for scalability, we consider a non-auto-regressive (NAR) scheme for parametrizing the probability distribution of topological orders. This allows our method to attain an extremely favorable performance vs. run time tradeoff: it always outperforms fast baselines, and is only matched or outperformed by those requiring a much longer (in one case 4000x more) run time.
    \item We address the problem of how to perform meaningful message-passing on DAGs, a graph type which has received comparatively less attention in the literature on Graph Neural Networks. We introduce \emph{Topoformer}, a flexible, attention-based architecture wherein messages can be passed between each and every pair of nodes, with a different set of learnable parameters depending on the topological relation between the nodes.
    \item To test our method, we introduce an algorithm for the generation of \emph{synthetic}, layered, Neural Net-like computation graphs, allowing any researcher to generate a dataset of \emph{as many as desired} graphs of \emph{any desired size}. These graphs are a more faithful model of real NN workflows, and allow us to prove our method on a much larger and varied dataset, than previous efforts~\cite{regalpaper}. To our knowledge, this is the first public algorithm of this kind. Nevertheless, we also test our method on proprietary graphs to illustrate its relevance to realistic compiler workflows.
    %\item Once a parametrization of the pdf of topological orders has been obtained through training of the aforementioned GNN, we implement and compare multiple ways of sampling sequences from it. We consider greedy sampling (i.e. choosing the most probable sequence, as per usual when deploying policy-based RL solutions), but also beam search (BS), as well as Beam Search with State Collapsing (BSC)%\corrado{Gotta cite references on BS here.}.
\end{itemize}

\section{Related work \label{sec:related}}

% non neural baselines: 
% main disandvantage of heuristics (eg LKH) is that time consuming to construct, do not transfer to other costs, instead our method black box optimizer -- we show it works for both costs functions.
% wrt simulated annealing/genetic algorithms we can fine tune for problem instances (but then they might ask about adaptive SA or hyperparams tuning for GA...)

% neural bselines: TSP papers (supervised vs RL, hybrid vs end2end).
% Could claim we get to larger size with RL and end2end than previous literature.
% SOP baseline: ...

% \herke{Adding some references here that might be useful:}
% On learning for routing (papers that either specifically adress more complex constriants, or AR vs NAR):

\paragraph{Machine Learning for Combinatorial Optimization:}
Combinatorial optimization as a use case for deep learning poses interesting technical challenges. First, the combinatorial nature of the problem conflicts with the differentiable structure of modern deep neural networks; and second, the models need to be run at large scale to solve real world instances, exacerbating the challenges in training deep learning models. Given the discrete nature of CO problems, a natural approach is to pose them as reinforcement learning (RL) problems~\cite{rl_book}. The aim is then to learn a policy that selects the best actions to maximize a reward directly related to the optimization objective. Algorithms then differ in the way the policy is parameterized: either in an end-to-end manner where the actions directly correspond to solutions of the optimization problem~\cite{gopaper, kool2018TSP, joshi2021learning, khalil2017learning}, or in a hybrid manner, where the policy augments parts of a traditional solver, e.g. by replacing heuristics used in setting parameters of an algorithm, see e.g. \cite{regalpaper, correia2022neural, xin2021neurolkh, ahn2020ordering}. Our approach follows an end-to-end design philosophy, which, not having to rely on an external algorithm, affords better control of post-compile run time and facilitates application on edge devices~\cite{ahn2020ordering}. Furthermore, RL has the advantage of being useful as a black box optimizer, when no handcrafted heuristics can be designed.\\
\textbf{Sequence Optimization via ML:} Within MLCO, much effort has been devoted to the task of predicting optimal sequences~\cite{bello2016neural,kool2018TSP,mena2018learning,linderman2018,gadetsky2020low}. The end-to-end nature of our method places it close to the one proposed in \cite{kool2018TSP}, although to the best of our knowledge, our work is the first to tackle the challenge of enforcing precedence constraints in the network predictions. As we shall see in more detail below, this generalization is non-trivial: already counting the number of topological orders belongs to the hardest class of computational problems \cite{sharp_p_complete}. This has to be contrasted with the fact that the number of sequences without topological constraints is simply $n!$ for $n$ objects.
Besides, as pointed out in \cite{joshi2021learning}, no MLCO method has so far been able to convincingly tackle TSPs of sizes above a few hundred nodes, when it comes to \emph{zero-shot} generalization to unseen problem instances, i.e.~when no fine tuning on the test set is done. It is also therein pointed out how an auto-regressive parametrization of the sequence (which was the method used in ref.~\cite{kool2018TSP}) appears to be necessary to achieve acceptable performance even at those small sizes. Conversely, in the present work we show compelling zero-shot performance on DAGs of sizes up to \emph{thousands} of nodes, while nonetheless generating our sequences in a fully non-auto-regressive (NAR) way and maintaining a strong run time advantage over classical sequencing algorithms. Our results can then also be interpreted as cautioning against the idea of using the TSP as the sole, paradigmatic test-bed for MLCO research, as \cite{joshi2021learning} remarks.\\
\textbf{ML for Compiler Optimization:}
The DAG sequencing task we consider is an omnipresent stage in compiler workflows, which usually also include such tasks as device assignment and operations fusion~\cite{gopaper}. In such a setting, jointly optimizing these tasks to reduce the \emph{run time} of a certain workflow (such as the forward pass of a Neural Net) is a common objective, which in refs~\cite{regalpaper, gopaper, steiner2021value} is tackled with ML methods. %However, this can make it hard to assess the merit of the solution on the individual tasks \roberto{don't they do separate tasks as well in GO?}, or render necessary the use of proprietary simulators to obtain the run time of both the ML solutions and the baselines it is compared with~\cite{gopaper}, which makes the results impossible to reproduce or build up on. 
In this work we focus on the task of minimizing the peak local memory usage during execution, which does not require a performance model or simulator as well as being relevant to applications on edge devices~\cite{ahn2020ordering}. In ~\cite{regalpaper}, the ML solution leans on an existing genetic algorithm, whilst our solution is end-to-end, much like that proposed in ~\cite{gopaper}. Another characteristic of the solution proposed in~\cite{gopaper} is the idea of interpolating between AR and NAR via an \emph{iterative refinement} scheme, in which sequences are generated in one pass but subsequently refined during an user-defined number of subsequent passes; conversely, we generate all our sequences in a single pass.\\
While in \cite{regalpaper} the run time optimization is studied on both real-world and synthetic random graphs -- the latter being relatively small (up to about 200 nodes), the peak memory optimization is studied only on a proprietary dataset augmented via perturbation of the node attributes.
In \cite{gopaper} the authors train and test their method on a relatively small set of six proprietary workflows which are not disclosed to the reader, and out of those six, only the size of the largest instance is mentioned.\\
%In our work, we conversely show compelling zero-shot generalization performance on a much larger and varied dataset (as large as desired by the user, in fact) of synthetic graphs of sizes of up to thousands of nodes, and which are a more realistic model of computation graphs typically encountered in NN compiler workflows (for example, they have well-defined layers and skip connections), with both their topology and node attributes varying between instances. We release the algorithm for generation of these graphs, hoping that it might become a standard in MLCO research addressing compiler workflows and sequence optimization in general. In addition, we report performance on a set of proprietary graphs.\\
%Finally, both~\cite{regalpaper} and~\cite{gopaper} make use of off-the-shelf NN architectures to parametrize their respective policies, whilst we propose and use a novel NN architecture for ours.
\textbf{Deep Graph Neural Networks:}
Given that our problem is specified as a DAG, it is a logical choice to parametrize our sequence-generation policy with a Graph Neural Network architecture~\cite{gnn_survey}. The basic idea of every GNN architecture is to update graph and edge representations by passing messages between the graph nodes along the graph edges~\cite{battaglia2018relational}. However, this can be too restrictive when it comes to sequence generation on DAGs. For example, nodes that come after each other in the sequence might not be linked by an edge in the graph, and therefore are unable to directly influence each other's representation. Notice how this difficulty is another consequence of the presence of precedence constraints in our problem, which conversely was not an issue in e.g.~\cite{kool2018TSP} where the graph is fully connected and no constraints are present.
Relatively few efforts (see e.g.~\cite{thost2020directed, zhang2019dvae, bianchini2002recursive}) have been devoted to devise a way to perform meaningful message passing on DAGs. As a matter of fact, the quest for expressive GNN architectures is at the center of intense theoretical investigation \cite{graphomer, review-expressive-gnn}.

\section{Background \label{sec:backg}}

\subsection{Topological Orders and DAGs}
We here introduce the mathematical background, starting with a few definitions. A partial order is an irreflexive transitive relation $<$ between certain pairs of a set $V$.
We call a pair $(x,y)\in V\times V$  that is related by $<$ comparable, and \emph{incomparable} otherwise.
A Directed Acyclic Graph (DAG) $G=(V,E)$ is a directed graph with no directed loops.
We can map a DAG $G=(V,E)$ to a partially ordered set $(V,<)$ where $x<y$ if there is a directed path from node $x$ to node $y$.
Multiple DAGs map to the same partial order. 
For example, the DAGs with vertex set $\{x,y,z\}$ and edge sets $E=\{x\to y, y\to z\}$ and
$E'=\{x\to y, y\to z, x\to z\}$, where $s\to t$ denotes a directed edge from $s$ to $t$, correspond to the same partial order $x<y<z$.
We define the \emph{transitive closure} (TC) of a DAG as the graph with most edges that has the same underlying partial order, so that there exists a directed edge $(x,y)$ whenever $x<y$. Conversely, the \emph{transitive reduction} (TR) is the graph with \emph{least} edges that results in the same partial order.
We denote the order induced by a DAG by $<_G$.

A topological order or sorting of a DAG $G$ is a bijection $\sigma : V \to \{1,\dots,|V|\}$ such that $\sigma(x)<\sigma(y)$ whenever $x<_Gy$. The set ${\cal T}_G$ of topological orders of $G$ is a subset of the permutation group of the vertices and coincides with total orders on $V$ that respect $<_G$, called \emph{linear extensions} of the partial order.
While there are several well-known algorithms to compute a topological order of a DAG, e.g. breadth first search and depth first search, counting the number of topological orders is one of the hardest computational problems, being $\#$P complete \cite{sharp_p_complete}.
%Efficient randomized algorithms based on Markov Chain Monte Carlo are however known to %approximate this number \cite{bubley1999faster}.
In this work we develop a general machine learning method to find a topological order that minimizes a given cost function on a DAG, which we define in the next section.

\subsection{Peak Memory Minimization}

Deciding the best way to schedule operations in a computational graph representing a neural network is a central problem in compilers \cite{regalpaper, gopaper, ahn2020ordering}.
We can associate a DAG to a computational graph in such a way that nodes represent operations ("ops"), and incoming/outgoing edges represent operands/results of these operations.
Every time one executes an op, the inputs\footnote{We use "inputs" and "operands" interchangeably throughout the paper.} to that op need to be in memory, and memory for the outputs needs to be allocated. Therefore, each node of the DAG carries a label 
$m : V \to \mathbb{N}$ specifying 
the memory required to store the output of that op.
A typical first step in scheduling a DAG is to identify topological orders to execute operations. 
% For example, the popular list scheduling algorithm requires a priority list which corresponds to a topological order of the ops to be executed by multiple processors \cite{}
% \roberto{to be checked and add ref}.
Compilers for edge devices, which have limited memory, aim at choosing the optimal topological order that minimizes the peak memory footprint \cite{ahn2020ordering}. %\mukul{I think the above paragraph should be part of introduction and motivation}
We focus therefore on the peak local memory usage minimization task, which can be formulated as the following combinatorial optimization problem on a labeled DAG $G=(V,E,m)$:
\begin{equation}
    \min_{\sigma \in {\cal T}_G} \C(\sigma),\qquad \C(\sigma) =
    \max(M_1(\sigma), \dots, M_{|V|}(\sigma)),
\end{equation}
with the definitions 
\begin{align}
    M_t &= I_{t-1} + m(\sigma_t) \label{eq:mem_cost},\\
    I_t &= M_t - \sum_{i \in S_t}m_i, \qquad  S_t = \left\{i: i \notin \bigcup_{l=0}^{t-1} S_{l} ~~ \text{ and }~~ \forall (i,j) \in E,\, j\in \sigma_{1:t} \right\}  \label{eq:I_t},
\end{align}
i.e. the memory usage at time $t$ is given by the memory usage $I_{t-1}$ of the outputs which have not yet been consumed, at time $t-1$, by downstream operations, plus the memory requirement of the output of operation $\sigma(t)$. $I_t$ is in turn obtained by subtracting from $M_t$ the memory costs of nodes whose outgoing edges only connect to already scheduled nodes, i.e. nodes whose output was only required by already scheduled operations. Naturally, $I_0 = 0, S_0 = \phi$. 
%\corrado{Q for Harris: how does the param cost fit into this?}\harris{params can be represented as coming from zero-indegree nodes with output size equal to the tensor size, so it fits without modification.  If we want to be specific, we could write ``input data and parameter data can be represented by zero-indegree `data' nodes in the graph whose output sizes equal their respective data sizes"}

% \subsection{Sequential Ordering Problem}

% The sequential ordering problem (SOP) is a variant of the celebrated traveling salesman (TSP) problem with precedence constraints. %\harris{this description is good but is "euclidian" only.  many SOP problems are not} 
% Formally, the TSP is defined by a set of cities $V$ together with a distance function $C:V\times V\to [0,\infty)$.
% The goal is to find the shortest tour, namely the permutation of the cities $\sigma \in S_V$ such that the total distance is minimal. The TSP is a well known NP-hard problem \cite{co_book}.
% In SOP, the problem is supplemented with precedence constraints on city pairs wherein any tour must visit city A before B, for example.  Such constraints can be represented by a DAG $G=(V,R)$ that restricts the search space of the TSP to topological orders of $G$.  Thus the solution of the SOP solves:
% \begin{align}
%     \min_{\sigma \in {\cal T}_G}
%     C_{\sigma(1),\sigma(2)}+ 
%     C_{\sigma(3),\sigma(3)}+ 
%     \dots 
%     +
%     %C_{\sigma(|V|-1),\sigma(|V|)}+ 
%     C_{\sigma(|V|),\sigma(1)}
%     \,.
% \end{align}
% Since the TSP is a particular case of the SOP, the SOP is NP-hard as well.

\section{Method}\label{sec:method}

\begin{figure}[htb!]
\centering
\includegraphics[width=0.95\linewidth]{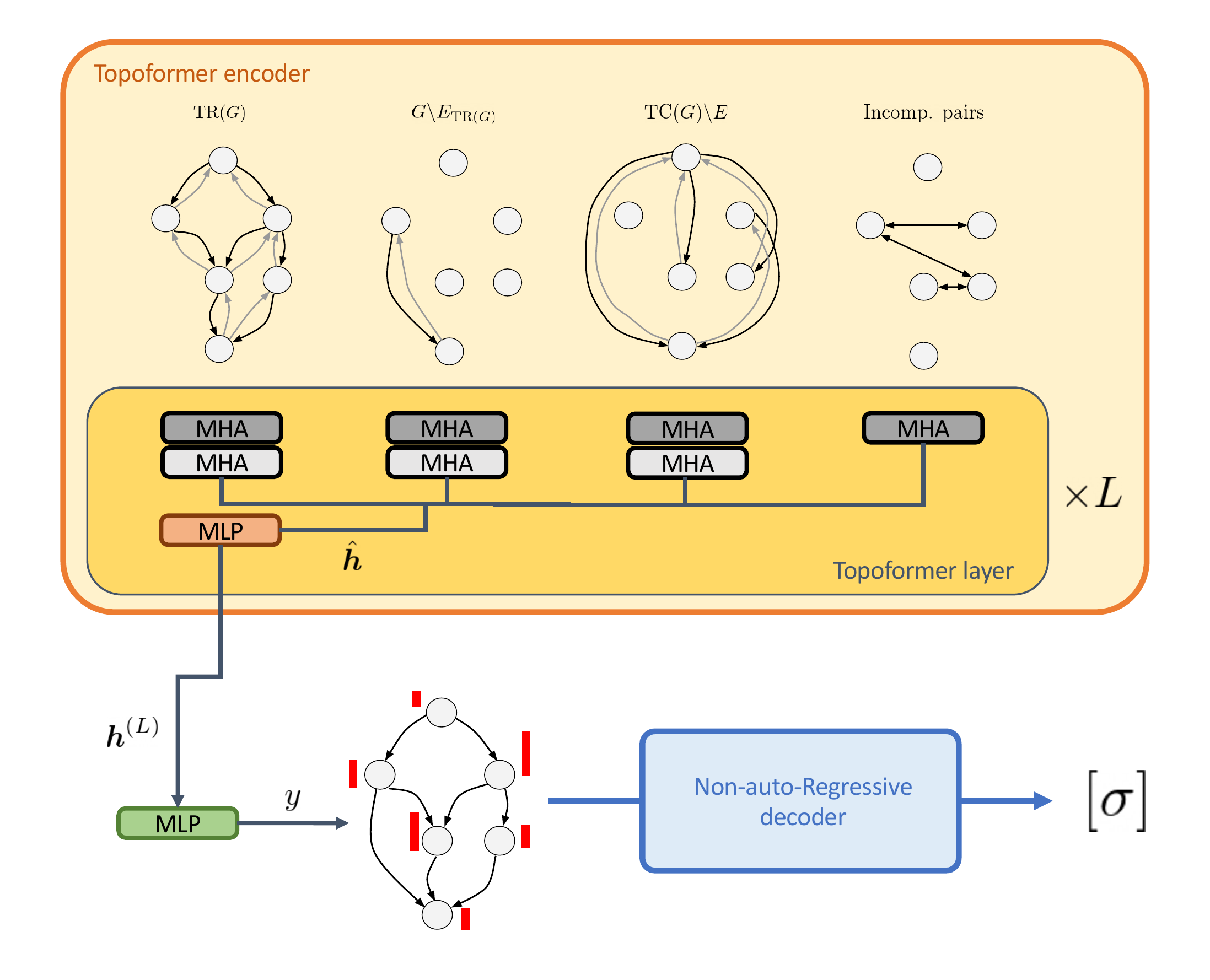}
\caption{Our complete architecture for neural topological ordering. The shades of gray in the MHA boxes are to highlight how attentions heads operate separately on the forward and backward version of the first three graphs. 
%\roberto{can we make the color of the boxes match those of the edges?
%Should the $\times L$ include also the different graphs since they are computed at each %layer?
%}
The priorities $(y_i)_{i=1}^{|V|}$ are represented by the red bars on the original DAG and decoded into a sequence with its associated probability. \label{fig:e2e_pipeline}}
\end{figure}

We use an encoder-decoder architecture whose schematic is shown in figure \ref{fig:e2e_pipeline}. Our encoder is \emph{Topoformer}, a novel GNN architecture, which derives an embedding for each node of the graph. The embeddings are used by the decoder which generates a distribution in the sequence space and finally the distribution can be converted to a sequence via different inference methods like sampling, greedy inference or beam search. Next, we describe each of the component in detail.

\subsection{Topoformer: Topologically Masked Attention\label{sec:topoformer}}

A Graph Neural Network (GNN) is a natural choice to encode our scheduling problem via embedding of the DAG nodes. All canonical GNN architectures operate by updating these embeddings via the aggregation of "messages" sent from the other nodes, usually in the form of some function of their own embedding~\cite{gnn_survey}. Architectures mainly differ in how the set of sender nodes is constructed and the aggregation function is chosen. In a Graph Convolutional Network~\cite{kipf2017semi}, the senders are the first neighbors of a node and the aggregation function is a weighted average, whilst in a vanilla Graph Attention Network~\cite{GAT}, the senders are all the other nodes, but their contributions are aggregated via averaging with \emph{learned} weights so as to account for their degree of relevance. When trying to apply such mechanisms on DAGs, a common point of contention is whether, and how in practice, the partial ordering encoded by it should reflect in the direction of travel of the messages~\cite{wang2021bilevel, bianchini2002recursive, thost2020directed}. While disregarding the DAG structure entirely (as one would do in a vanilla GAT), does not appear wise, it might be too restrictive when it comes to our task. For example, nodes that are next to each other in the sequence might well be incomparable, and thus lack a path for messages between them. The combinatorial nature of the task also poses requirements; it is known~\cite{graphomer, bengio2021machine} that reasoning about CO problems on a graph requires the capacity to reason about the \emph{global structure} of it, whilst architectures such as those proposed in~\cite{wang2021bilevel, bianchini2002recursive, thost2020directed} limit the set of sender nodes to a \emph{local} neighborhood of the receiver node. In summary, our architecture must strike a compromise between accounting for \emph{global} structure and \emph{local} partial ordering information.

Our \emph{Topoformer} architecture meets these requirements. A vector $\bx_i$ of input features (see the appendix for details about its definition and dimensionality) is first turned into an initial node embedding $\bh_i^{(0)}$ via a node-wise linear transformation, $\bh_i^{(0)} = W\bx_i + \bbias$. Subsequently, a succession of $L$ attention layers, each of them consisting of a Multi-Head Attention (MHA)~\cite{GAT} sub-layer followed by one more node-wise MLP, updates these embeddings, similar to a vanilla Transformer~\cite{vaswani2017attention}; however, we confer a topological inductive bias to these updates by having a separate group of attention heads masked by each of the following graphs induced by the original DAG:
\begin{itemize}[leftmargin=4.2mm,noitemsep]
    \item Its transitive reduction (TR).
    \item The directed graph obtained by removing the TR edges from the DAG: $G\backslash E_{\textrm{TR}(G)}$.
    \item The directed graph obtained by removing the edges of the DAG from its TC: $\textrm{TC}(G)\backslash E$.
    \item The backwards versions (i.e. with flipped edges) of each of the three above.
    \item The undirected graph obtained by joining all incomparable node pairs.
\end{itemize}
By adding together these graphs, one would obtain the fully connected graph relative to the node set $V$, whereupon all nodes would attend to all nodes. Then effectively, the propagation rules of Topoformer are same as those of a vanilla transformer encoder,
\begin{align}
    \hat{\bh}_i^{(\ell)} =& \bh_{i}^{(\ell-1)} + \texttt{concat}_{j}\left[\MHA_i^{\ell, j}\left(\bh_1^{(\ell-1)},\dots,\bh_{|V|}^{(\ell-1)}; M^j\right)\right], \\
    \bh^{(\ell)}_i=& \hat{\bh}_i^{(\ell)} + \MLP^{(\ell)}\left(\hat{\bh}_i^{(\ell)}\right) \label{eq:mlptopo},
\end{align}
save for the presence of the \emph{mask} $M^j$, which ensures that head $j$ only attends to its assigned graph among the seven listed above. Following~\cite{transf_layernorm}, we also apply layer normalization~\cite{layernorm} to the MHA and MLP inputs. The number of heads assigned to each graph can be chosen independently (setting it to zero means to not message-pass along the edges of the respective graph), or parameters can be tied among different MHAs. One should also remark how the MLP sub-layer allows the flow of information between different attention heads. All nodes are then able to influence each other's representation, while anyway injecting a strong inductive bias based on the DAG structure. Information about the Topoformer configurations used in our experiments is provided in the appendix.

\subsection{Decoder} \label{sec:ar_nar}
Once the embeddings of the nodes are generated, the decoder's task is to derive a stochastic policy $p(\sigma|G)$ over the valid topological orders of the graph. The most straightforward way is to take advantage of the chain rule of conditional probability to decompose the policy as a product
\beq
p(\sigma|G) = \prod_{t=2}^{|V|}p_\theta(\sigma_t|\sigma_{1:t-1},\bh , G) \times p_\theta(\sigma_1|\bh , G) ,
\eeq
We could then sample a complete sequence by autoregressively choosing a new node at each step as done e.g. in~\cite{kool2018TSP}. This scheme is the most principled and expressive; however, when a NN is used as a function approximator for $p_\theta$, it also requires that $|V|$ calls to this NN be performed, which limits its feasibility to relatively small graphs due to the amount of computation required.

%This makes it acceptable to sacrifice expressivity for run time, by employing a Non-Auto-Regressive (NAR) scheme which decouples the number of NN calls from the graph size.
In order to scale to large graphs, we employ a Non-Auto-Regressive (NAR) scheme which decouples the number of NN calls from the graph size.
Similar to the approach of~\cite{gopaper}, we assign scheduling \emph{priorities} $y_i
\in \mathbb{R}$ to the nodes, rather than scheduling probabilities. 
The priority for node $i$ is derived by passing its final embedding through an MLP:
\beq
y_i = \textrm{MLP}\left(\bh^{(L)}_i\right). \label{eq: mlp_priority}
\eeq
These priorities are assigned with a \emph{single} NN inference. The sequence itself is subsequently constructed by adding a new node at each step. Given the partial sequence $\sigma_{1:i-1}$, the next node can only be selected from a subset $\mathcal{S}(\sigma_{1:i-1}, G)$ of schedulable nodes, due to both the graph topology and choices made earlier in the sequence. Then, the distribution of the next node to be added at step $i$ is given as follows:
\beq
p(\sigma_t|\sigma_{1:t-1}, \bh, G) = \begin{cases} \dfrac{\exp(y_{\sigma_t})}{\sum_{j\in \mathcal{S}(\sigma_{1:t-1}, G)} \exp(y_j)}, ~~ &\text{if } \sigma_t \in \mathcal{S}(\sigma_{1:t-1}, G), \\
0, ~~~ &\text{otherwise.}
\end{cases}
\eeq

%\roberto{Change with: (note fixed a few typos)
%\beq
%\label{eq:nar_probs}
%p(\sigma_i|\sigma_{1:i-1}, y, G) = \frac{\exp(y_{\sigma_i})}{\sum_{j\in %\mathcal{S}(\sigma_{1:i-1}, G)} \exp(y_j)} I_{\mathcal{S}(\sigma_{1:i-1}, G)}(\sigma_i)
%\,,
%\eeq
%with $I_A(x)=1$ if  $x\in A$ and $0$ otherwise.
%Recall now the Plackett-Luce distribution over a sequence %$\sigma=(\sigma_1,\dots,\sigma_N)$:
%\begin{equation}
%    p_{\text{PL}}(\sigma)
%    =
%    \prod_{i=1}^N
%    \frac{\exp(y_{\sigma_i})}
%    {\sum_{j=i}^N \exp(y_{\sigma_j})}
%    =
%    \prod_{i=1}^N
%    \frac{\exp(y_{\sigma_i})}
%    {\sum_{j\in \mathcal{S}(\sigma_{1:i-1}, K_N)} \exp(y_j)}
%    \,,
%\end{equation}
%where we denoted by $\mathcal{S}(\sigma_{1:i-1}, K_N)$ the set of nodes in the complete graph $K_N$ excluding those corresponding to $\sigma_{1:i-1}$. 
%Comparing against \eqref{eq:nar_probs} we see that the probability distribution induced by our NAR model provides a natural topological generalization of the Plackett-Luce distribution to sequences with precedence constraints. 
%In particular, our decoder can approximate a Dirac distribution over topological orders that has mass concentrated on the solution to the peak memory minimization problem for a given DAG $G$ -- see \cite{gadetsky2020low} for a related statement in the case of sequences without precedence constraints.
%In this sense, we do not lose expressive power for solving combinatorial optimization problems on DAGs by using the NAR model instead of an AR model.
%}

\textbf{Decoding Methods}: 
We use the following three methods to obtain the next node in the partial sequence from the distribution $p(\sigma_t|\sigma_{1:t-1}, \bh, G)$:
\begin{enumerate}[leftmargin=4.5mm]
    \item \emph{Greedy}: At each step $t$, select the node with the highest probability i.e. $\sigma_t = \argmax_{\Tilde{\sigma}_t} p(\Tilde{\sigma}_t|\sigma_{1:t-1}, \bh, G)$
    \item \emph{Sampling}: At each step $t$, sample from the next node distribution i.e. $\sigma_t \sim p(\cdot|\sigma_{1:t-1}, \bh, G)$
    \item \emph{Beam search with state-collapsing}: We can also expand the partial sequences by using a beam search method where the score function is total probability of the partial sequence. We improve our beam search routine by making the following observation: suppose there are two partial sequences in consideration, $\sigma_{1:t}$ and $\Tilde{\sigma}_{1:t}$, such that both have scheduled the same set of nodes so far (but different order), and $\C(\sigma_{1:t}) < \C(\Tilde{\sigma}_{1:t})$. Then, we can ignore the partial sequence $\Tilde{\sigma}_{1:t}$ and only keep $\sigma_{1:t}$ in the beam search. This is because both partial sequences must schedule the same set of remaining nodes, and hence the set of future memory costs are identical for both $\sigma_{1:t}$ and $\Tilde{\sigma}_{1:t}$, but the current peak memory cost is higher for $\Tilde{\sigma}_{1:t}$. Thus, $\sigma_{1:t}$ dominates $\Tilde{\sigma}_{1:t}$ in terms of achievable minimal peak memory usage.
\end{enumerate}

\subsection{Training}

Our encoder-decoder architecture induces a distribution $p_{\theta} (\sigma|G)$ on the set of topological orders for a given DAG $G$. The expected cost incurred is given by $J(\theta | G) = \mathbb{E}_{p_\theta(\sigma|G)}\left[\mathcal{C}(\sigma(\theta))\right]$. We minimize the cost $J(\theta) = \mathbb{E}_{G} \left[J(\theta | G)\right]$ via gradient descent using the REINFORCE gradient estimator ~\cite{williams1992,rl_book} as follows 
\begin{equation}
    \nabla J(\theta) = \mathbb{E}_{G, p_\theta(\sigma|G)}\left[(\mathcal{C}(\sigma) - b(G)\right)\nabla_\theta\log p_\theta(\sigma|G)],
\end{equation}
where $b(G)$ is a \emph{baseline} meant to reduce the variance of the estimator. We follow~\cite{kool2018TSP} in setting it equal to the cost of a \emph{greedy rollout} of a baseline policy on the graph $G$
\begin{equation}
    b(G) = \C(\argmax_\sigma p_\theta(\sigma|G)).
\end{equation}

\section{Experiments}\label{sec:experiments}

We conduct experiments on a synthetic dataset of graphs which we refer to as "layered graphs", as well as a set of real-world computation graphs. We compare our approach with the following classic topological ordering baselines: 
\begin{itemize}[leftmargin=3.2mm,noitemsep]
     \item\emph{Depth/Breadth first sequencing}: Find the topological order by traversing the graph in  depth/breadth first manner according to the layout of the graph generated using pygraphviz.
     \item\emph{Depth-first dynamic programming (DP)}: Depth-first DP is a global depth-first method for searching the optimal sequence, with automatic backtracking when equivalent partial sequences are found; it retains the full sequence with minimum cost so far, and returns it if search does not complete before the prescribed timeout.
     \item\emph{Approximate DP}: In approximate DP, a beam of partial sequences are considered at each step. For each beam in the subsequent step, the top-$K$ partial sequences with the lowest costs are retained. This DP is also able to find the optimal sequence given enough memory and compute resources (with unlimited beam size $K$), but we only consider an approximate version with $K=10^5$ in this work. Note that approximate DP uses the state-collapsing and parallelism to improve its efficiency.
     
     %we consider the state space being the set of nodes that have been selected so far and the action space being a set of schedulable nodes on a given state. With such a model, we observed that DP can be used to find out the \emph{exact} solution if we have sufficient memory and computational resources. Due to the combinatorial characteristics of our problem, we used approximated version of DP.
     \item\emph{Random order}: We generate $100$ random topological orders, and pick the one with smallest cost.
\end{itemize}
Please see the appendix for more detailed description of the baselines. Therein, we also report some ablation studies on various neural baselines including ablation studies on decoder by considering other neural architectures, as well as a comparison with an end to end baseline adapted from ref.~\cite{kool2018TSP}. Neural topo order greedy, sample and BS denote the performance of our model in greedy, sampling and beam search inference mode respectively. We use a sample size and beam size of $16$ sequences, of which the best one is subsequently picked, for all our experiments. Next, we describe in detail the results of the two experiments.

\subsection{Layered Graphs}

In order to generate a large corpus of training data we come up with a way to synthetically generate graphs of a given size which have similar structure to the computation graphs of feed-forward neural networks. We call our synthetic graph family \emph{layered graphs}, as these graphs comprise of well-defined layers of nodes. The nodes in a layer have connections to the nodes in the subsequent layer and can also have skip connections with nodes in layers farther down. The number of layers, number of node per layer, number of edges between subsequent layers, number of skip connections and memory utilization of the nodes are all generated randomly, and can be controlled by setting appropriate parameters. We refer the reader to the appendix for more details on layered graphs, including their generation algorithm and some visual examples.

We train our model on $500$-node layered graphs for $325$ epochs, where in each epoch we generate a training set of $1000$ new graphs. We test the performance of our model on a set of 300 unseen graphs of the same size, generated with the same method. We also evaluate the cross-size generalization performance of our trained model by testing it on graphs of size $1000$ and $2000$. We refer the reader to the appendix for a comprehensive description of the training algorithm and model configuration.

\begin{figure}[htb!]
\centering
% \begin{subfigure}{.33\textwidth}
%   \centering
%   \includegraphics[width=0.99\linewidth]{figures/500_0.14_speed_vs_perf.png}
%   \label{fig:tcm_100}
% \end{subfigure}%
% \begin{subfigure}{.33\textwidth}
%   \centering
%   \includegraphics[width=0.99\linewidth]{figures/1000_0.14_speed_vs_perf.png}
%   \label{fig:tcm_500}
% \end{subfigure}
% \begin{subfigure}{.33\textwidth}
%   \centering
%   \includegraphics[width=0.99\linewidth]{figures/2000_0.14_speed_vs_perf.png}
%   \label{fig:tcm_2k}
% \end{subfigure}
\includegraphics[width=\linewidth]{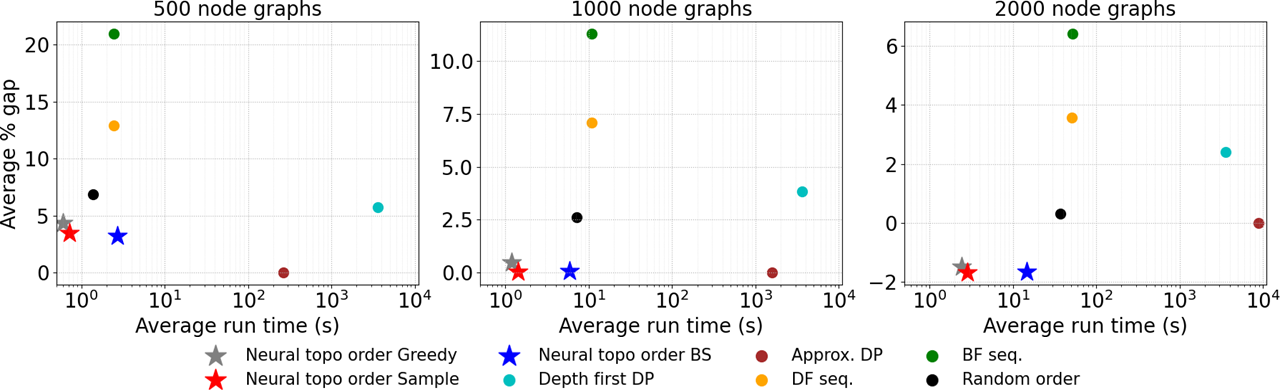}
\caption{Average \% gap from approximate DP vs average run time comparison on the test set of 300 layered graphs. Lower is better for both \% gap and run time.
\label{fig:perf_time_layered}}
\end{figure}

Figure \ref{fig:perf_time_layered} shows the performance vs. run time plot on layered graphs of size $|V|=500, 1000$, and $2000$. We report the performance in terms of the \% gap of peak memory utilization from the peak memory obtained via approximate DP, which we consistently observed to be the best-performing baseline. Note that the run time is plotted on a log-scale. We can observe that for $500$-node graphs, our model beats all the baselines except approximate DP in terms of both the memory usage and run time. Our model is slightly worse than approximate DP from the memory usage perspective, but it runs 100x faster. We also observe that our model generalizes well to larger sized graphs. For the case of $2000$-node graphs our model performs better than approximate DP in terms of peak memory usage, while being $4000$x times faster. This shows that while approximate DP performs more poorly as graph size increases, our model is able to generalize to larger graphs by learning meaningful embeddings of the topological structure thanks to Topoformer, and to be extremely fast thanks to our NAR decoding scheme.
\begin{table}[h!tb]
\begin{adjustwidth}{-0.25cm}{}
    \centering
    \caption{Comparison of methods on the synthetic layered graph test set.}
    \vspace{0.3cm}
    \begin{center}
    \begin{tabular}{lrrrrrr}
        \toprule
        
        \multicolumn{1}{c}{\multirow{3}{*}{Algorithm}} & 
        \multicolumn{2}{c}{{500-    node graphs}} & 
        \multicolumn{2}{c}{{1000-node graphs}} &
        \multicolumn{2}{c}{{2000-node graphs}}  \\
        \cmidrule{2-3}  \cmidrule{4-5} \cmidrule{6-7}
        & \% gap from   & run time & \% gap from   & run time & \% gap from   & run time  \\
        & approx. DP   & [$s$] & approx. DP   & [$s$] & approx. DP & [$s$] \\
        \midrule
        Approximated DP & 0 & 264.88 & 0 & 1561.17  & 0 & 8828.86  \\
        \midrule
        Depth-First DP & 5.76 & 3600 & 3.84 & 3600 & 2.40 & 3600 \\
        (max. run time=1H) & & & & &  &  \\
        Random order & 6.86 & 1.38 & 2.62 & 7.13 & 0.31 & 36.87 \\
        Depth-first seq. & 12.9 & 2.45 & 7.1 & 10.91 & 3.57 & 51.32 \\
        Breadth-first seq. & 20.94 & 2.43 & 11.31 & 10.87 & 6.42 & 51.52 \\
        \midrule
        Neural Topo Order & & & & & & \\
        \checkmark Greedy & 4.32 & 0.6 & 0.48 & 1.19 & -1.47 & 2.44 \\
        \checkmark Sample & 3.49 & 0.72 & \textbf{0.03} & 1.41 & \textbf{-1.68} & 2.87 \\
        \checkmark Beam search & \textbf{3.21} & 2.68 & 0.08 & 5.92 & -1.66 & 14.74 \\
        \bottomrule
    \end{tabular}
    \end{center}
    \label{tab:decoding_complexity}
\end{adjustwidth}
\end{table}

\subsection{Real-World Graphs}

\begin{figure}[htb!]
\centering
\includegraphics[width=0.5\textwidth]{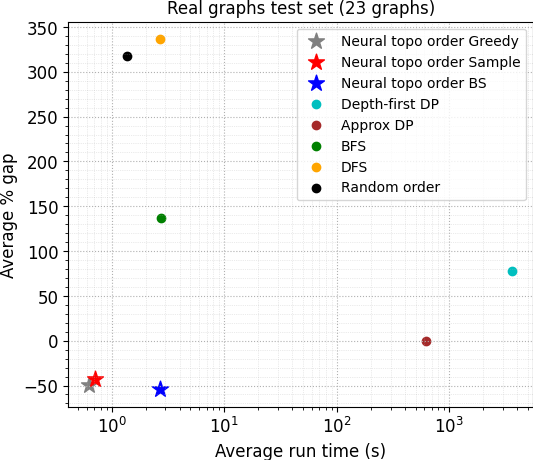}
\caption{Performance vs run time comparison for different approaches on test set of real computation graphs. Performance is measured in average \% gap from approximate DP.\label{fig:perf_time_real}}
\end{figure}

While our synthetic layered graphs are convenient for experimentation, we see value in also presenting results obtained from neural computation graphs used for commercial development of our artificial intelligence hardware and software products.  Here we sample 115 representative graphs that have diverse architectures (classifiers, language processors, denoisers, etc.) and size (from a few dozen to 1k nodes). We split this dataset into a training set and test set via a random $80-20$ split. We train our model for $500$ epochs and report the performance on the unseen test set at the end of training in table \ref{tab:real_graph}. In order to ensure fair comparison of run times, we stratify the test set into 3 categories based on the graph size.  
Figure \ref{fig:perf_time_real} shows the performance vs run time plot on the test set of real graphs. We observe that for real graphs the performance gap between the best baseline (approximate DP) and our model is remarkable. We can obtain sequences which are $50\%$ better than approximate DP on average while also being almost 1000x faster on average. This proves the capability of our model to generalize and perform well on real-world computation workflows. 

\begin{table}[h!tb]
\begin{adjustwidth}{-0.25cm}{}
    \centering
    \caption{Comparison of methods on the real graph test set. Smaller \% gap is better}
    \vspace{0.3cm}
    \begin{tabular}{lrrrrrr}
        \toprule
        
        \multicolumn{1}{c}{\multirow{3}{*}{Algorithm}} & 
        \multicolumn{2}{c}{{200 - 500-node graphs}} & 
        \multicolumn{2}{c}{{500 - 700-node graphs}} &
        \multicolumn{2}{c}{{700 - 1000-node graphs}}  \\
        \cmidrule{2-3}  \cmidrule{4-5} \cmidrule{6-7}
        & \% gap from   & run time & \% gap from   & run time & \% gap from   & run time  \\
        & approx. DP   & [$s$] & approx. DP   & [$s$] & approx. DP & [$s$] \\
        \midrule
        Approximated DP & 0 & 113.54 & 0 & 517.60  & 0 & 1131.61  \\
        \midrule
        Depth-First DP & 62.18 & 3600 & 102.76 & 3600 & 50.57 & 3600 \\
        (max. run time=1H) & & & & & & \\
        Random order & 469.34 & 0.25 & 376.16 & 1.24 & 116.24 & 2.40 \\
        Depth-first seq. & 506.21 & 0.70 & 394.93 & 2.26 & 123.21 & 4.49 \\
        Breadth-first seq. & 348.77 & 0.75 & 149.81 & 2.31 & -35.55 & 4.86 \\
        \midrule
        Neural Topo Order & & & & & & \\
        \checkmark Greedy & -17.57 & 0.42 & -51.23 & 0.6 & -68.97 & 0.83 \\
        \checkmark Sample & \textbf{-21.53} & 0.44 & -40.51 & 0.68 & -61.46 & 0.97 \\
        \checkmark Beam search & -19.5 & 1.22 & \textbf{-57.34} & 2.58 & \textbf{-73.45} & 3.86 \\
        \bottomrule
    \end{tabular}
    \label{tab:real_graph}
\end{adjustwidth}
\end{table}

\subsection{Encoder Ablation Study}
 
\begin{table}[h!tb]
\begin{adjustwidth}{-0.25cm}{}
    \centering
    \caption{Comparison of different encoder architectures. Topoformer with MP (message passing) on DAG corresponds to forward and backward message passing only on the input DAG using topoformer.}
    \vspace{0.3cm}
    \begin{center}
    \begin{tabular}{lrrrrrr}
        \toprule
        
        \multicolumn{1}{c}{\multirow{3}{*}{Algorithm}} & 
        \multicolumn{2}{c}{{500-node graphs}} & 
        \multicolumn{2}{c}{{1000-node graphs}}\\
        \cmidrule{2-3}  \cmidrule{4-5} 
        & \% gap from   & run time & \% gap from   & run time  \\
        & approx. DP   & [$s$] & approx. DP   & [$s$] \\
        \midrule
        MLP & & & & \\
        \checkmark Greedy & 8.31 $\pm$ 0.76 & 0.58 $\pm$ 0.0 & 2.95 $\pm$ 0.48 & 1.52 $\pm$ 0.01 \\
        \checkmark Sample & 4.41 $\pm$ 0.50 & 0.67 $\pm$ 0.0 & 0.68 $\pm$ 0.35 & 1.84 $\pm$ 0.02 \\
        \checkmark Beam search & 6.5 $\pm$ 0.69 & 2.47 $\pm$ 0.01 & 2.43 $\pm$ 0.49 & 7.62 $\pm$ 0.07 \\
        \midrule
        Fully Connected Transformer & & & & \\
        \checkmark Greedy & 8.46 $\pm$ 0.72 & 0.69 $\pm$ 0.01 & 3.09 $\pm$ 0.46 & 1.3 $\pm$ 0.01 \\
        \checkmark Sample & 4.72 $\pm$ 0.52 & 0.8 $\pm$ 0.01 & 0.85 $\pm$ 0.37 & 1.55 $\pm$ 0.02 \\
        \checkmark Beam search & 6.52 $\pm$ 0.72 & 2.98 $\pm$ 0.03 & 2.09 $\pm$ 0.47 & 6.49 $\pm$ 0.07 \\
         \midrule
         GAT (forward only) & & & & \\
         \checkmark Greedy & 5.94 $\pm$ 0.61 & 0.49 $\pm$ 0.01 & 1.33 $\pm$ 0.38 & 1.24 $\pm$ 0.01 \\
        \checkmark Sample & 4.19 $\pm$ 0.56 & 0.64 $\pm$ 0.01 & 0.48 $\pm$ 0.36 & 1.54 $\pm$ 0.02 \\
         \checkmark Beam search & 4.22 $\pm$ 0.60 & 2.22 $\pm$ 0.02 & 0.60 $\pm$ 0.38 & 5.94 $\pm$ 0.04 \\
        \midrule
         GAT (forward+backward) & & & & \\
         \checkmark Greedy & 4.84 $\pm$ 0.55 & 0.63 $\pm$ 0.01 & 0.90 $\pm$ 0.37 & 1.37 $\pm$ 0.02 \\
         \checkmark Sample & 3.55 $\pm$ 0.53 & 0.80 $\pm$ 0.01 & 0.23 $\pm$ 0.36 & 1.67 $\pm$ 0.02 \\
         \checkmark Beam search & 3.55 $\pm$ 0.54 & 2.89 $\pm$ 0.01 & 0.39 $\pm$ 0.36 & 6.50 $\pm$ 0.05 \\
        \midrule
        Topoformer (forward+backward) (Ours) & & & & \\
        \checkmark Greedy & 4.82 $\pm$ 0.55 & 0.73 $\pm$ 0.01 & 0.76 $\pm$ 0.36 & 1.62 $\pm$ 0.02  \\
        \checkmark Sample & 3.67 $\pm$ 0.52 & 0.85 $\pm$ 0.01 & 0.21 $\pm$ 0.36 & 1.99 $\pm$ 0.02 \\
        \checkmark Beam search & 3.68 $\pm$ 0.57 & 3.03 $\pm$ 0.03 & 0.35 $\pm$ 0.37 & 8.1 $\pm$ 0.08 \\
        \midrule
        Full Topoformer (Ours) & & & & \\
        \checkmark Greedy & 4.31 $\pm$ 0.56 & 1.04 $\pm$ 0.01 & 0.47 $\pm$ 0.36 & 1.51 $\pm$ 0.01  \\
        \checkmark Sample & 3.35 $\pm$ 0.52 & 1.21 $\pm$ 0.01 & \textbf{-0.01} $\pm$ 0.35 & 1.8 $\pm$ 0.02 \\
        \checkmark Beam search & \textbf{3.08} $\pm$ 0.51 & 4.15 $\pm$ 0.02 & 0.05 $\pm$ 0.36 & 7.4 $\pm$ 0.07 \\
        \bottomrule
    \end{tabular}
    \end{center}
    \label{tab:encoder_ablation}
\end{adjustwidth}
\end{table}

We conduct experiments by using an MLP, fully connected transformer and GAT as an encoder architecture to quantify the effectiveness of our topoformer architecture. We test a vanilla version of GAT (referred as GAT forward only) which does message passing only on the edges of the DAG. We also consider GAT encoder which does message passing on the augmented graph having reverse edges corresponding to all the edges of the DAG and refer to this setting as GAT forward+backward.

We train each model on the layered graph dataset of 500 node graphs. We evaluate the performance of the trained model on the test set (300 graphs) of 500 node and 1000 node graphs. We use a sample size and beam width of 16 for evaluation on both 500 and 1000 node graphs. The MLP and transformer use the same number of layers and hidden dimension as the topoformer specified in appendix \ref{append:architecture_details}. We run the inference on our test set of $300$ graphs $10$ times for each model to be more precise in our run time calculations. We report the mean \% gap from approximate DP and the mean run time across all the graphs and trials along with their 95\% confidence interval. 
 
Table \ref{tab:encoder_ablation} shows the performance of different encoder architectures. It can be observed that both versions of our topoformer architecture and GAT have a superior performance than MLP and fully connected transformer for both graph sizes. Moreover, full topoformer (message passing on all the seven graphs listed in section \ref{sec:topoformer}) has a better performance than GAT and topoformer with message passing only the forward and backward edges of the DAG. This shows the benefit of global message passing between all the nodes which is enabled by the full topoformer.

\section{Conclusion}

In this work we propose an end-to-end machine learning method for the task of optimizing topological orders in a directed acyclic graph. Two key elements in our design are: (1) an attention-based GNN architecture named Topoformer that employs message passing that is both global and topologically-aware in directed acyclic graphs, (2) a non-autoregressive parametrization of the distribution on topological orders that enables fast inference. We demonstrated, for both synthetic and real-world graphs, the effectiveness of the method in tackling the problem of minimizing peak local memory usage for a compute graph -- a canonical task in compiler pipelines. Said pipelines also include other tasks~\cite{gopaper}, chief amongst them the one of assigning operations to devices for execution. At the present stage, our method and dataset cannot be leveraged for solving these, or for end-to-end optimization of a whole pipeline. Extending our method to this more challenging setting is therefore a natural direction for future research.
%We believe that our proposed method opens up possibilities of applying ML to a large class of topological ordering related combinatorial optimization problems, and it paved the way for solving more challenging problems in graph compilation such as multi-device scheduling of compute graphs.

\bibliographystyle{IEEEtran}
\bibliography{refs}

%% Commented out for arxiv version
%\input{checklist}
%%%%%%%%%%%%%%%%%%%%%%%%%%%%%%%%%%%%%%%%%%%%%%%%%%%%%%%%%%%%

\appendix

%\section{Appendix}
\newpage
%\faketableofcontents
\doparttoc % Tell to minitoc to generate a toc for the parts
\addcontentsline{toc}{section}{Appendix} % Add the appendix text to the document TOC
\part{Appendix} % Start the appendix part
\parttoc % Insert the appendix TOC

\section{Layered graphs dataset}
We report here the details of the generation algorithm we use to create our dataset.  It is not the first time that a synthetic dataset of graphs is used to train and test an ML framework on a compiler task, as this was already done in ref.~\cite{regalpaper}. However, the models therein used were generic random graph models (e.g. Erdos-Renyi), rather than a model explicitly tailored to reproduce NN-like computation graphs. We develop such a model, and we release its details with the intent of both ensuring reproducibility of our results, as well as of providing tool that we hope will be picked up by researchers interested in compiler problems, as well as more general sequence optimization task on DAGs.

The algorithm builds a graph by organizing a fixed number $|V|$ of nodes into well-defined layers, and then placing edges between subsequent layers, as well as skip connections that skip at least one layer. While the number of nodes is fixed by the user, the target number of layers $L$ depends on the \emph{width factor} $\mathcal{W}$ of the graph. A width factor of 0 would result in a one-dimensional chain graph, whilst a width factor of 1 in a graph with a single, wide layer,
\beq
L = \left\lceil\sqrt{|V|\left(\frac{1}{\mathcal{W}}-1\right)}\right\rceil,
\eeq
where $\lceil\cdot\rceil$ is the ceiling function. In order to promote architectural variability within the dataset, we choose to randomly draw a new width factor, $\mathcal{W}\sim U\left(0.25,0.5\right)$, for each graph, with $U(a,b)$ denoting the uniform distribution in the $[a,b]$ interval. Subsequently, the number of nodes to assign to each layer $\ell$ is also an integer randomly drawn from a uniform distribution
\beq
\Nl \sim U\left(\lceil|V|/L\left(1-\sigma_\mathcal{N}\right)\rceil,\lfloor|V|/L\left(1+\sigma_\mathcal{N}\right)\rfloor\right),
\eeq
with $\sigma_\mathcal{N}$ being a user-defined variability parameter, and $\lfloor\cdot\rfloor$ is the floor function. We stress that both $L$ and $\Nl$ are just target values, since we wish to keep $|V|$ fixed: this layer-by-layer node addition process is stopped as soon as the graph has the number of nodes $|V|$ required, which might lead to the number of layers and nodes per layer being ultimately different from their respective targets. The pseudocode for this procedure is reported in algorithm~\ref{algo:nodes}.
\begin{algorithm}[htb!]
\SetAlgoLined
\KwOut{A layered graph $G=(V,)$ without edges}
\KwIn{Total number of nodes $|V|$, number-of-nodes-per-layer variability $\sigma_\mathcal{N}$}
\KwData{layer index $\ell$, node index $n$, node counter $N$, target number $\N_\ell$ of nodes for layer $\ell$}
$\ell \leftarrow$ 0\;
$N \leftarrow$ 0\;
\While{True}{
% \uIf{$N\geq |V|$}{
% break\;
% }
$\Nl \sim U\left(\lceil|V|/L\left(1-\sigma_\mathcal{N}\right)\rceil,\lfloor|V|/L\left(1+\sigma_\mathcal{N}\rfloor\right)\right)$\;
\For{$n\in[1, \Nl]$}{
\uIf{$N\geq |V|$}{
break\;
}
add node $n$ to graph $G$\;
add node $n$ to layer $\ell$\;
$N \leftarrow N+1$\;
}
$\ell \leftarrow \ell + 1$
}
\caption{Node-assignment algorithm for layered graphs. \label{algo:nodes}}
\end{algorithm}

After the layers are set up, the algorithm proceeds to assign edges between adjacent layers. As an example, let us assume that $\mathcal{N}_1$ and $\mathcal{N}_2$ are the numbers of nodes for two adjacent layers, with $\N_2<\N_1$. The maximal number of edges between these two layers, corresponding to a fully-connected, MLP-like topology, would be $\N_1\times\N_2$. Since we want each node to have at least one ingoing and one outgoing connection (except for those in the first and last layers), the minimal number of connections must be $\max(\N_1,\N_2) = \N_1$. The user can interpolate between these two extrema by tuning the \emph{edge density} parameter $\rho_E$, with the number of edges to place between the two layers being ultimately equal to
\beq
|E|_{(\ell_i,\ell_{i+1})} = (\N_{\ell_i}\times\N_{\ell_{i+1}})\rho_E + (1-\rho_E)\max(\N_{\ell_i},\N_{\ell_{i+1}}).
\eeq
This budget of edges is subsequently distributed among the nodes in the larger layer (layer 1 in our example), with them being assigned to the node with the smallest number of so-far-assigned edges (ties are broken randomly), until it is exhausted.
What then remains to do is connecting all the so assigned edges to nodes in the other layer (layer 2 in our example above). We choose these destination nodes in a such a way that, if the layers were visualized as being centered one above the other, with the larger layer at the top, the edges assigned to a node end up more or less equally spaced in 2-$d$ cone below it. This procedure is repeated for every pair of adjacent layers, as we report in algorithm~\ref{algo:edges}.
\begin{algorithm}[htb!]
\SetAlgoLined
\KwOut{A layered graph $G=(V,E)$ with edges but no skip connections.}
\KwIn{A layered graph $G=(V,)$ without edges, edge density $\rho_E$}
\caption{Edge-assignment algorithm for layered graphs. \label{algo:edges}}
\KwData{Number $|E|_{(\ell_i,\ell_j)}$ of edges between layers $\ell_i$ and $\ell_j$. $c_n$ is a counter of edges incoming or outgoing from node $n$}
\For{$\ell_1 \in$ graph layers}{
$\ell_2 = \ell_1+1$\;
$|E|_{(\ell_1,\ell_2)} = (\N_{\ell_1}\times\N_{\ell_2})\rho_E + (1-\rho_E)\max(\N_{\ell_1},\N_{\ell_2})$ (rounded to the closest integer)\;
\uIf{$\N_1\geq\N_2$}{
$\ell_s \leftarrow \ell_1$, $\ell_t \leftarrow \ell_2$\;
}
\Else{ 
$\ell_s \leftarrow \ell_2$, $\ell_t \leftarrow \ell_1$\;
}
\For{$n \in \ell_s$}{
$c_n \leftarrow 0$\;
}
\While{$\sum_{n\in \ell_s}c_n < |E|_{(\ell_1,\ell_2)}$}{
$\mathcal{S} \leftarrow \argmin c_n$\;
Pick $i$ randomly from set $\mathcal{S}$\;
$c_i \leftarrow c_i + 1$\;

}
\For{$n \in [0, \mathcal{N}_{\ell_s}-1]$}{
\uIf{$\N_{\ell_s}=1$}{
$n_c$ = 0\;
}\Else{
$n_c = n \times \frac{\N_{\ell_t}-1}{{\N_{\ell_s}}-1}$ set "center node," rounded to the nearest integer
}
\For{$i \in[0,c_n-1]$}{
%$n_t = n_c - c_n//2 + i$ (if $n_t < 1$, shift all $n_t$s up so that the lowest $n_t$ is equal to 1\;
$n_t = (n_c - (c_n - 1)//2) + [0, c_n-1]$ (a range centered at $n_c$)\;
Shift the range $n_t$ up/down such that no index is less than $0$ or greater than $\N_{\ell_t}-1$ \;
\For{$j \in n_t$}{
add one edge between node $n$ of layer $\ell_s$ and node $j$ of layer $\ell_t$
}
% \uIf{$\N_{\ell_1}\leq \N_{\ell_2}$}{
% add one edge between node $n_t$ of layer $\ell_1$ and node $n$ of layer $\ell_2$\;
% }\Else{
% add one edge between node $n$ of layer $\ell_1$ and node $n_t$ of layer $\ell_2$\;
%}
}
}
}
\end{algorithm}

\begin{figure}[htb!]
\centering
\begin{subfigure}{.33\textwidth}
  \centering
  \includegraphics[width=\textwidth]{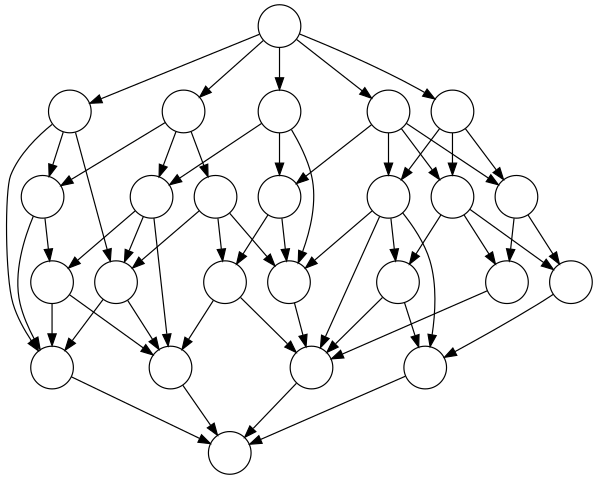}
  %\label{fig:rd_kodak}
\end{subfigure}%
\begin{subfigure}{.33\textwidth}
  \centering
  \includegraphics[width=\textwidth]{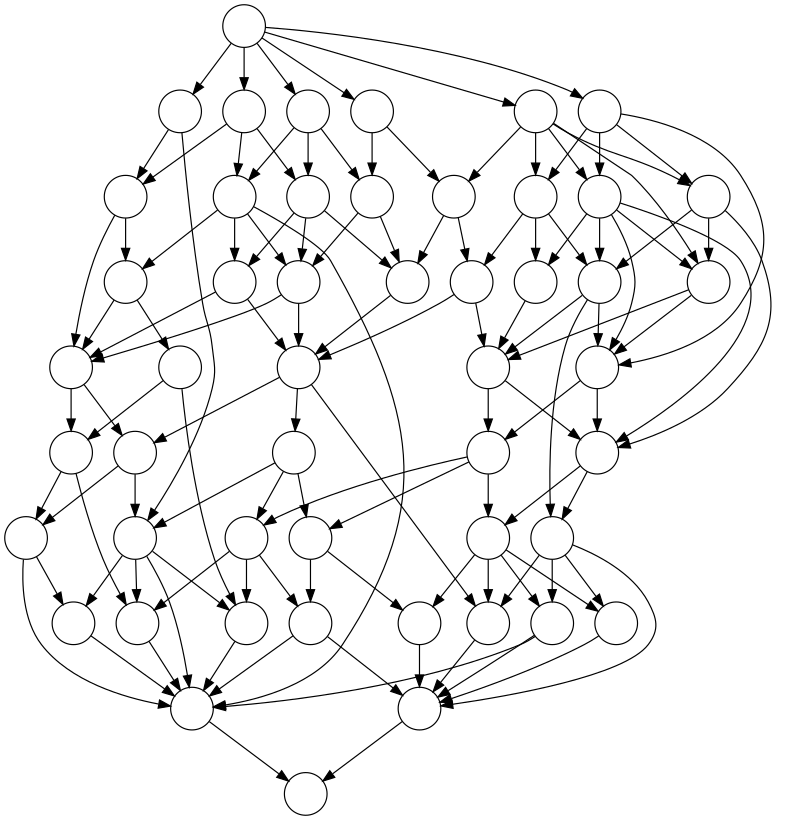}
  %\label{fig:rd_kodak}
\end{subfigure}%
\begin{subfigure}{.33\textwidth}
  \centering
  \includegraphics[width=\textwidth]{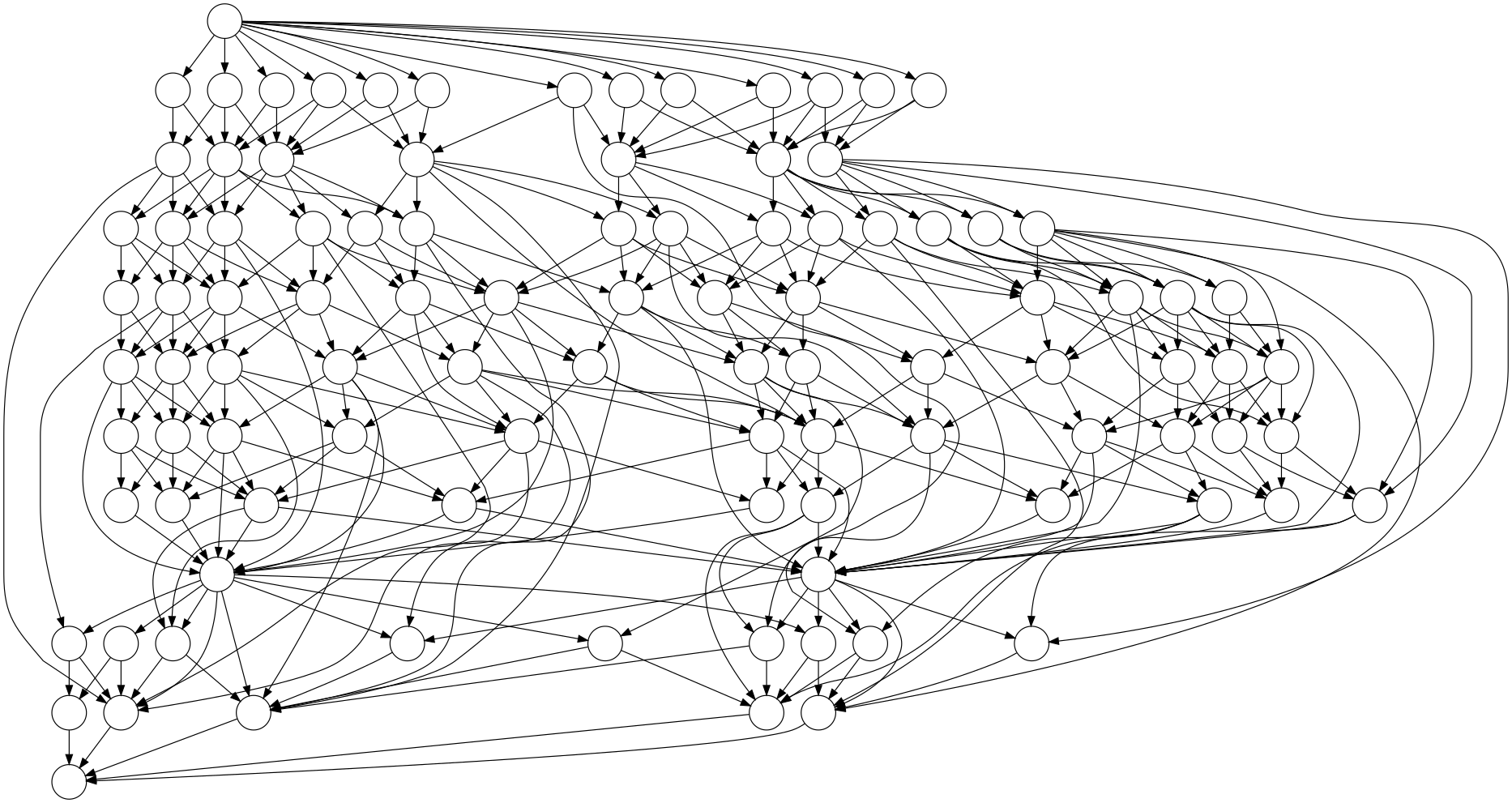}
  %\label{fig:rd_kodak}
\end{subfigure}%
\caption{Three example graphs from the layered graph family with (from left) 25, 50, and 100 nodes, generated using the algorithm we describe in the text. One can clearly make out the layered structure and easily remark the presence of skip connections.\label{fig:layered_graph}}
\end{figure}

Skip connections, i.e. edges skipping at least one layer, which are often found in modern NN architectures, are then added to the graph. The total number of skip connections to add is fixed as
\beq
\N_S = |E|\frac{\rho_S}{(1 - \rho_S)},
\eeq
where $|E|$ is the total number of edges in the graph so far, and $\rho_S$ a user-defined skip connection density. For each skip connection, we randomly draw a source layer among those between the first and the third-to-final ones (since skip connections must skip at least one layer). The target layer number is then also drawn at random between the source layer number $+2$, and the final layer (both included). One must then assign a source and a target \emph{node} within each of these layers. We just select the source node at random within the source layer, and then assign the target node in such a way that it would be more or less directly below the source node if the graph were visualized on a 2-$d$ plane. The pseudocode of this procedure is reported in algorithm~\ref{algo:skip}, and figure \ref{fig:layered_graph} shows three example instances of layered graphs created with our algorithm.
\begin{algorithm}[htb!]
\SetAlgoLined
\KwOut{A layered graph $G=(V,E)$ with both connections between adjacent layers, and skip connections}
\KwIn{A layered graph $G=(V,E)$ with edges between adjacent layers but no skip connections, skip connection density $\rho_S$}
\KwData{number of layers $L$, number of edges $|E|$}
\uIf{L<3}{
break; ~~\tcc{cannot have skip connections with fewer than 3 layers}\
}
$\N_S = \lceil|E|\frac{\rho_S}{(1 - \rho_S)}\rceil$\;
\For{$i\in [0,\N_S)$}{
$\ell_s \leftarrow$ a layer at random between the first and third-to-last (both included)\;
$\ell_t \leftarrow$ a layer at random between layer number $\ell_s + 2$ and the last (both included)\;
$x_s\sim U(0,1)$\;
$y\sim U(0,1)$\;
$x_t = x_s + 0.2\times y$\;
$x_t = \min(x_t, 0.999)$; ~~\tcc{ensure that $x_t\in[0,1)$}\  
%$x_t = \max(x_t, 0)$\;
% add an edge between node $\lfloor x_s (\N_{\ell_s}+1)\rfloor$ of layer $\ell_s$ and node $\lfloor x_t (\N_{\ell_t}+1)\rfloor$ of layer $\ell_t$
add an edge between node $\lfloor x_s \N_{\ell_s}\rfloor$ of layer $\ell_s$ and node $\lfloor x_t \N_{\ell_t}\rfloor$ of layer $\ell_t$
}
\caption{Skip connection-assignment algorithm for layered graphs. \label{algo:skip}}
\end{algorithm}

Finally, we specify the assignment of memory costs to the nodes. In the layered graph model, we have both output memory costs $\left(m_i\right)_{i=1}^{|V|}$ and parameter costs $\left(p_i\right)_{i=1}^{|V|}$, where the output cost is the memory usage of the output of an operation, and the parameter cost the one of a variable necessary to execute the operation; for example, if the operation at node $i$ were a matrix multiplication, $\boldsymbol{y}=M\boldsymbol{x}$, $o_i$ would be the memory usage of $\boldsymbol{y}$ and $p_i$ the one of the matrix $M$. The parameter cost of operation $\sigma_t$ during a sequence is added to the memory usage at time $t$, but not to the cost at subsequent steps since the memory associated to it can be de-allocated as soon as the operation has been executed.In particular, the memory utilization cost $M_t$ in \eqref{eq:mem_cost} gets modified to the following:
\beq
 M_t = I_{t-1} + m(\sigma_t) + p(\sigma_t)
\eeq
where $I_t$ is defined in \eqref{eq:I_t}. Both costs are randomly drawn from a simple mixture of Gaussians $\text{GMM}(\mathbf{w}, \mathbf{\mu}, \mathbf{\sigma}) \equiv \sum_{i=1}^4 w_i\N(\mu_i,\sigma_i)$ projected on to the positive reals,
% \beq
% \rho(o) = \rho(p) \equiv \sum_{i=1}^4 w_i\N(o;m_i,\sigma_i).
% \eeq
\beq
m_i \sim \text{GMM}(\mathbf{w}, \mathbf{\mu}, \mathbf{\sigma})\big|_{\mathbb{R}_+}, ~~~~ p_i \sim \text{GMM}(\mathbf{w}, \mathbf{\mu}, \mathbf{\sigma}))\big|_{\mathbb{R}_+}.
\eeq

To align the costs assignment with the real world computation graphs, instead of sampling the memory costs for each node $n$, we sample one output cost $m_l$ and parameter cost $p_l$ for each layer $l$ and assign the costs $m_l, p_l$ to each node in layer $l$. This is because many real world computation graphs are a tiled version of the original precedence graph of compute nodes where each node is broken down into a layer of nodes with similar shape and parameter requirements. 
This concludes the description of our dataset generation algorithm. For the sake of reproducibility, we report below the value we took for all the user-defined parameters mentioned in this section:
\begin{itemize}
\item Variability of number of nodes per layer $\sigma_\N = 0.75$
\item Edge density $\rho_E = 0.2$
\item Skip connection density $\rho_S = 0.14$
\item Means of the Gaussian mixture $(\mu_1,\mu_2,\mu_3,\mu_4) = (0.5, 1, 3, 5)$
\item Standard deviations of the Gaussian mixture $(\sigma_1,\sigma_2,\sigma_3, \sigma_4) = (0.5, 1, 1, 1)$
\item Weights of the Gaussian mixture $(w_1,w_2,w_3,w_4) = (0.3, 0.3, 0.3, 0.1)$
\end{itemize}

\section{Decoder Ablation study}

% In order to measure the effectiveness of our architecture we perform ablation experiments to study the effect of changing the encoder (Table~\ref{tab:encoder_ablation}), changing the decoder to an auto-regressive decoder and changing both the encoder and the decoder (Table~\ref{tab:decoder_ablation}). 

In order to measure the effectiveness of our architecture we perform ablation experiments to study the effect of changing the decoder to an auto-regressive decoder and changing both the encoder and the decoder (Table~\ref{tab:decoder_ablation}).

We compare the performance of our architecture with the model which uses topoformer as an encoder but uses an auto-regressive decoder. We adapt the decoder designed for the TSP problem \cite{kool2018TSP} for our memory-minimization problem. The decoder of \cite{kool2018TSP} uses a notion of context node for decoding and at each decoding step using a series of multi-head attention with the context node arrives at the distribution of the next node to be selected for the order. We modify the masking procedure in the decoder of \cite{kool2018TSP} to mask out all the nodes which are not present in the set of feasible next nodes $\mathcal{S}(\sigma_{1:t-1}, G)$.

 We also conduct an experiment by changing both the encoder and decoder by adapting the model of \cite{kool2018TSP} to our problem. We adapt the auto-regressive decoder of \cite{kool2018TSP} as described above.  \cite{kool2018TSP} uses a fully connected transformer as an encoder since the underlying graph in TSP is a fully connected graph. We modify the encoder of \cite{kool2018TSP} to do message passing only on the edges of our input DAG so that it can exploit the topological structure of the graph in the encoding stage. We refer to this model as "GNN encoder + AR decoder" in table \ref{tab:decoder_ablation}. 
 
 \begin{table}[h!tb]
\begin{adjustwidth}{-0.25cm}{}
    \centering
    \caption{Comparison with Auto-regressive decoding}
    \vspace{0.3cm}
    \begin{center}
    \begin{tabular}{lrrrrrr}
        \toprule
        
        \multicolumn{1}{c}{\multirow{3}{*}{Algorithm}} & 
        \multicolumn{2}{c}{{500-node graphs}} & 
        \multicolumn{2}{c}{{1000-node graphs}}\\
        \cmidrule{2-3}  \cmidrule{4-5} 
        & \% gap from   & run time & \% gap from   & run time  \\
        & approx. DP   & [$s$] & approx. DP   & [$s$] \\
        \midrule
        GNN encoder + AR decoder & & & & \\
        \checkmark Greedy & 6.13 $\pm$ 0.58 & 1.66 $\pm$ 0.01 & 1.84 $\pm$ 0.39 & 3.34 $\pm$ 0.02 \\
        \checkmark Sample & 4.71 $\pm$ 0.56 & 1.76 $\pm$ 0.01 & 1.38 $\pm$ 0.37 & 3.59 $\pm$ 0.02 \\
        \checkmark Beam search & 4.87 $\pm$ 0.61 & 4.01 $\pm$ 0.02 & 2.09 $\pm$ 0.41 & 7.90 $\pm$ 0.05 \\
        \midrule
        Topoformer + AR decoder & & & & \\
        \checkmark Greedy & 4.43 $\pm$ 0.55 & 1.53 $\pm$ 0.01 & 0.53 $\pm$ 0.35 & 3.05 $\pm$ 0.02 \\
        \checkmark Sample & 3.33 $\pm$ 0.51 & 1.7 $\pm$ 0.01 & \textbf{0.05} $\pm$ 0.35 & 3.38 $\pm$ 0.02 \\
        \checkmark Beam search & 3.14 $\pm$ 0.52 & 4.27 $\pm$ 0.04 & 0.13 $\pm$ 0.36 & 7.90 $\pm$ 0.05 \\
        \midrule
        Topoformer + NAR decoder (Ours) & & & & \\
        \checkmark Greedy & 4.31 $\pm$ 0.56 & 1.04 $\pm$ 0.01 & 0.47 $\pm$ 0.36 & 1.53 $\pm$ 0.01  \\
        \checkmark Sample & 3.35 $\pm$ 0.52 & 1.21 $\pm$ 0.01 & 0.09 $\pm$ 0.35 & 1.78 $\pm$ 0.01\\
        \checkmark Beam search & \textbf{3.08} $\pm$ 0.51 & 4.15 $\pm$ 0.02 & 0.2 $\pm$ 0.36 & 5.57 $\pm$ 0.05 \\
        \bottomrule
    \end{tabular}
    \end{center}
    \label{tab:decoder_ablation}
\end{adjustwidth}
\end{table} 
 
 We train both the models: "GNN encoder + AR decoder" and "Topoformer + AR decoder" on the layered graph dataset of 500 node graphs. We evaluate the performance of the trained model on the test set (300 graphs) of 500 node and 1000 node graphs. We use a sample size and beam width of 16 for 500 node graphs and a sample size and beam width of 8 for 1000 node graphs. We use a smaller sample size for 1000 node graphs due to GPU memory issues with the auto-regressive decoder approaches.

 Table \ref{tab:decoder_ablation} shows the mean and the 95\% confidence interval of the \% gap from approximate DP and run time for the three approaches on 500 and 1000 node graphs. We note that the performance of topoformer with AR decoder is quite close to our model for both 500 and 1000 node graphs. However, our model can run inference 2x faster than topoformer with AR decoder on 1000 graphs nodes (in greedy mode). Also, our model outperforms the adaptation of \cite{kool2018TSP} attention based GNN encoder and AR decoder to our problem both in terms of memory cost of sequence and run time. This shows the merit of our topoformer architecture over using a traditional GNN architecture which does message passing only on the input graph.

% \begin{itemize}
%     \item Changing the Decoder: We compare the performance of our architecture with the model which uses topoformer as an encoder but uses an auto-regressive decoder. We adapt the decoder designed for the TSP problem \cite{kool2018TSP} for our memory-minimization problem. The decoder of \cite{kool2018TSP} uses a notion of context node for decoding and at each decoding step using a series of multi-head attention with the context node arrives at the distribution of the next node to be selected for the order. We modify the masking procedure in the decoder of \cite{kool2018TSP} to mask out all the nodes which are not present in the set of feasible next nodes $\mathcal{S}(\sigma_{1:t-1}, G)$.
%     \item Changing the Encoder: We conduct experiments by using an MLP and fully connected transformer as an encoder architecture to quantify the effectiveness of our topoformer architecture. We keep the decoder fixed to our non-autoregressive decoder for these experiments
%     \item Changing both Encoder and Decoder: We also conduct an experiment by changing both the encoder and decoder by adapting the model of \cite{kool2018TSP} to our problem. We adapt the auto-regressive decoder of \cite{kool2018TSP} as described above.  \cite{kool2018TSP} uses a fully connected transformer as an encoder since the underlying graph in TSP is a fully connected graph. We modify the encoder of \cite{kool2018TSP} to do message passing only on the edges of our input DAG so that it can exploit the topological structure of the graph in the encoding stage. 
% \end{itemize}

\section{Training and Model details}\label{append:architecture_details}

\subsection{Training}
We train our model using the ADAM optimizer with the initial learning rate of $10^{-4}$ and learning rate decay factor of $0.996$ per epoch. We use a batch size of $8$ for training our model. The training and testing of our model is done on a single GPU (Nvidia Tesla V-100) with $32$ GB memory. We trained our model for 326 epochs on the synthetic graph dataset where in each epoch we provide 1000 training graphs. Also, we provide a new training set in each epoch so that we do not overfit our model on a fixed training set. We found the training to be fairly stable, and it converged in about 1-2 days.

\subsection{Model architecture}
 We use topoformer with number of layers $n_{layers} = 4$, embedding dimension $d = 256$, number of heads $n_{heads} = 10$ for each MHA operation on the seven graphs listed in section \ref{sec:topoformer} and the query and value dimension of $64$ for each head of MHA. The MLP used in \eqref{eq:mlptopo} consists of a linear layer ($d_{input} = d_{output} = 256$) with GELU activation followed by another linear layer ($d_{input} = d_{output} = 256$). The MLP used in \eqref{eq: mlp_priority} to generate the node priorities consists of a linear layer ($d_{input} = d_{output} = 256$) with RELU activation followed by another linear layer ($d_{input} = 256, d_{output} = 1$). In order to restrict the range of priority values, we also normalize the priorities of the nodes used for the decoding as follows: 
\begin{equation}
    \Tilde{y}_i = \alpha \times \frac{y_i - \text{mean}(\mathbf{y})}{\text{std} (\mathbf{y})}
\end{equation}
where $\mathbf{y} = \left[y_1, y_2, \ldots, y_{|V|} \right]$ and $\alpha$ is a hyperparamter. We set $\alpha=5$ for our experiments. 

\subsection{Baselines}
We provide more details about the dynamic programming baselines used in our experiments to compare the performance of our model

\begin{itemize}
    
    \item 
    \textbf{Depth-First Dynamic Programming (DP).}
    Topological orders are generated in a depth-first manner (with backtracking) where next node is picked randomly among available candidates.  Branch exploration is terminated if 1) the same set of nodes are in the partial sequence as a branch that has been already explored - only the lowest cost partial sequence is retained (dynamic programming approach), and 2) if the current partial cost is already higher than the lowest cost of any full sequence already found (cost increases monotonically).   This algorithm will eventually find the global optimal order, though the run time for doing so is expected to be at least exponential in |V|~\cite{ahn2020ordering}; it is however able to return at least one complete sequence in time $O(|V| + |E|)$~\cite{kahn_dfs} in the worst case, same as DFS.
    In our implementation, we set a wall time of one hour and pick the best complete path found. We observe that for our synthetic layered graphs, if the graph size is as small as $|V|=100$, we can actually find the optimal sequence in most cases within the one hour budget. We ran this algorithm on a CPU machine with Intel(R) Xeon(R) W-2123 CPU @ 3.60GHz
    
    \item 
    \textbf{Approximate DP.}
    % Approximate Dynamic Programming: In this approach, a set (i.e.~a \emph{beam}) of partial sequences is considered at each step $t$, and \emph{all} of the possible next-node choices for each of these are examined; for each partial sequence, only the choice leading to the lowest memory usage at $t+1$ is selected, generating the beam for $t+1$. This algorithm would also theoretically able to find the optimal sequence, if one took a beam size equal to $|{\cal T}_G|$, the cardinality of the set of topological orders of $G$. This is obviously infeasible, so we designate it as \emph{approximate} DP and take a fixed beam size.
    We define the state space $S$ as the space including a set of all nodes for each partial sequence (which \emph{ignores} the ordering information) and the action space for each state as the space of all possible next-node choices at that state (based on the topological structure). As an example for the state representation, if there is a partial sequence $5\rightarrow 2 \rightarrow 4 \rightarrow 3 \rightarrow 1$, the corresponding state is $\{1, 2, 3, 4, 5\}$.
    With the empty set $\emptyset$ being an initial state (meaning that no node has been added), we consider a state transition model that adds an action (a node) to a state and creates a successor state. Specifically, we can partition $S$ into $S_0\cup S_1\cup \cdots \cup S_{|V|}$, where $S_t$ is the space including a set of all nodes for each length-$t$ partial sequence (note that $S_0=\{\emptyset\}$). At every iteration $t=0, 1, ..., |V|-1$, the algorithm takes $S_t$ and assumes that we have \emph{(1) the minimum cost} and \emph{(2) the best partial sequence} for each state in $S_t$, where the minimum cost is over all feasible partial sequences corresponding to the state. Then, for each successor state in $S_{t+1}$, the algorithm computes the minimum cost and the best partial sequence for reaching out that state. 
    
    It should be noted that the algorithm gives an \emph{exact} solution if the amount of time and memory resource is sufficient, e.g., an exact solution can be found for 100-node graphs. However, due to the practical resource limitation, we only keep top-$K$ elements of $S_{t+1}$ for each iteration $t$ based on costs. We use the beam size $K=100,000$ for all experiments, and Nvidia Tesla V-100 is used for parallel computation across multiple states for each iteration.

\end{itemize}

\subsection{Baseline policy}
 The baseline $b(G)$ used in the policy gradient update is generated using the greedy rollout of the baseline policy. The baseline policy is also an instance of our model which is updated regularly during the course of training. At the end of each epoch, if the performance of the model being trained becomes better than the baseline model (in greedy inference mode) on a set of validation graphs then we copy the weights of the trained mode to the baseline model.

\subsection{Input features and initial node embedding}  We use the following as the input features $x_j$ for node $j$:
\begin{enumerate}
    \item Output memory cost $m_j$ and parameter memory cost $p_j$
    \item In-degree and out-degree of the node
    \item Minimum and maximum distance (in terms of hop count) of the node from the source and target node
\end{enumerate}

We normalize each entry of the input node feature across the nodes so that the features lie between $0$ and $1$ making it invariant with respect to the graph size. To be precise, the $i^{th}$ entry of the normalized input feature of node $j$ is given as $\bar{x}^i_j = \frac{x^i_j}{\max_{n} x^i_n}$. We also augment the node features with the Laplacian positional encodings (PE) \cite{dwivedi2020generalization} of dimension 20. We compute the laplacian PE using the laplacian matrix of the undirected DAG where all the directed edges are converted to undirected edges. Finally, the initial embedding $h^0_j$ for node $j$ is obtained by passing $\bar{x}_j$ through a linear layer.

%In order to highlight the merits of Topoformer and NAR, we also compare with a neural baseline, which we obtain by adapting to our task the architecture used in~\cite{kool2018TSP}, which is composed of a vanilla Transformer encoder~\cite{vaswani2017attention} and an AR decoder.

\end{document}